\documentclass[sigconf]{acmart}

\usepackage{multirow}
\usepackage[normalem]{ulem}
\useunder{\uline}{\ul}{}

\AtBeginDocument{%
  \providecommand\BibTeX{{%
    \normalfont B\kern-0.5em{\scshape i\kern-0.25em b}\kern-0.8em\TeX}}}

\copyrightyear{2022}
\acmYear{2022}
\setcopyright{acmcopyright}
\acmConference[SIGIR '22]{Proceedings of the 45th International ACM SIGIR Conference on Research and Development in Information Retrieval}{July 11--15, 2022}{Madrid, Spain.}
\acmBooktitle{Proceedings of the 45th International ACM SIGIR Conference on Research and Development in Information Retrieval (SIGIR '22), July 11--15, 2022, Madrid, Spain}
\acmPrice{15.00}
\acmISBN{978-1-4503-8732-3/22/07}
\acmDOI{10.1145/XXXXXX.XXXXXX}

\begin{document}

\fancyhead{}

\title{Training Entire-Space Models for Target-oriented Opinion Words Extraction}

\author{Yuncong Li}
\affiliation{%
	\institution{Tencent Inc}
	\city{Shenzhen}
	\country{China}
}
\email{yuncongli@tencent.com}

\author{Fang Wang}
\affiliation{%
	\institution{Shenzhen University}
	\city{Shenzhen}
	\country{China}
}
\email{2160230414@email.szu.edu.cn}

\author{Sheng-Hua Zhong}
\authornote{Corresponding author}
\affiliation{%
	\institution{Shenzhen University}
	\city{Shenzhen}
	\country{China}
}
\email{csshzhong@szu.edu.cn}


\begin{abstract}
  Target-oriented opinion words extraction (TOWE) is a subtask of aspect-based sentiment analysis (ABSA). Given a sentence and an aspect term occurring in the sentence, TOWE extracts the corresponding opinion words for the aspect term. TOWE has two types of instance. In the first type, aspect terms are associated with at least one opinion word, while in the second type, aspect terms do not have corresponding opinion words. However, previous researches trained and evaluated their models with only the first type of instance, resulting in a sample selection bias problem. Specifically, TOWE models were trained with only the first type of instance, while these models would be utilized to make inference on the entire space with both the first type of instance and the second type of instance. Thus, the generalization performance will be hurt. Moreover, the performance of these models on the first type of instance cannot reflect their performance on entire space. To validate the sample selection bias problem, four popular TOWE datasets containing only aspect terms associated with at least one opinion word are extended and additionally include aspect terms without corresponding opinion words. Experimental results on these datasets show that training TOWE models on entire space will significantly improve model performance and evaluating TOWE models only on the first type of instance will overestimate model performance\footnote{Data and code are available at https://github.com/l294265421/SIGIR22-TOWE}. 
\end{abstract}





\maketitle

\section{Introduction}
\begin{figure}[t]
	\centering
	\includegraphics[width=0.9\linewidth]{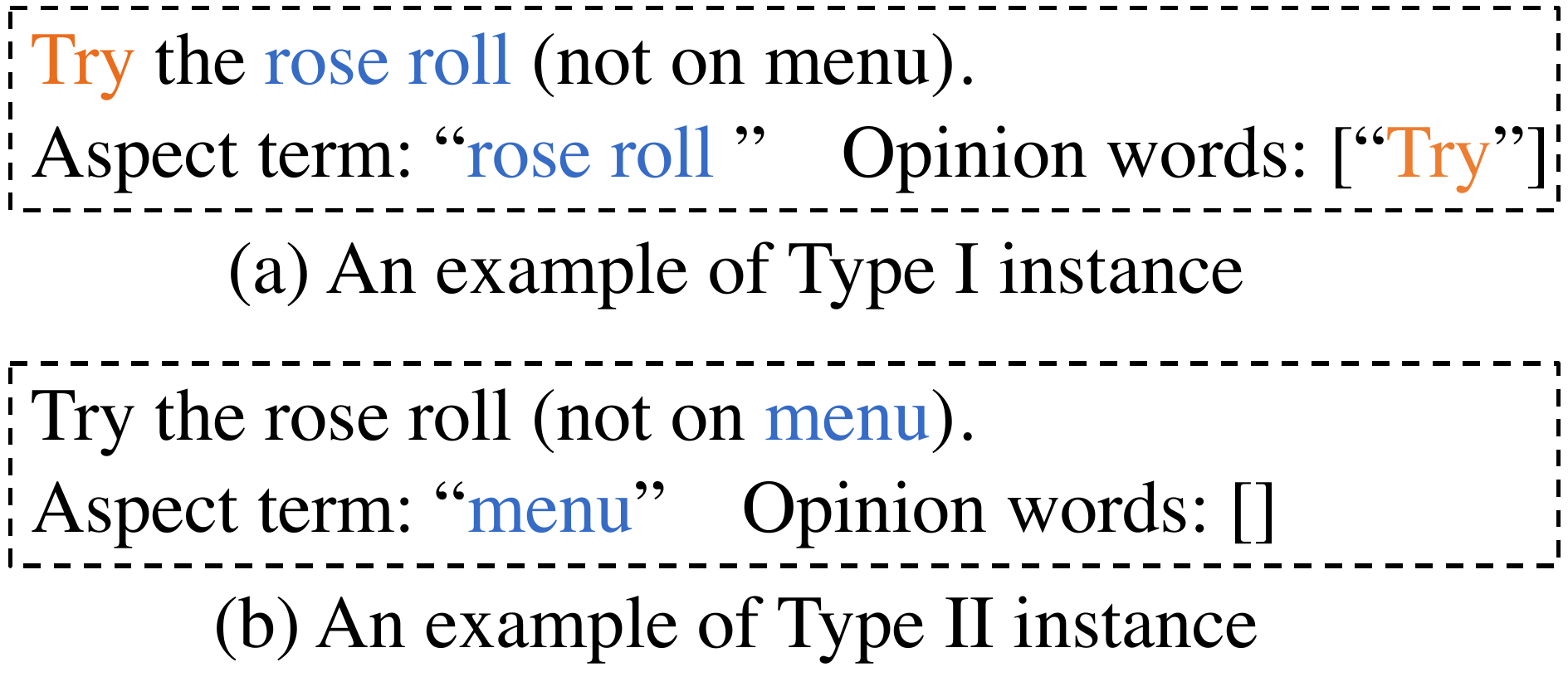}
	\caption{Two examples of TOWE task. The words highlighted in blue represent the given aspect terms, while the words in orange represent the corresponding opinion words.}
	\label{examples}
\end{figure}

Aspect-based sentiment analysis (ABSA) \cite{10.1145/1014052.1014073, pontiki-etal-2014-semeval, pontiki-etal-2015-semeval, pontiki-etal-2016-semeval} is a branch of sentiment analysis~\citep{10.1145/945645.945658, liu2012sentiment}. Target-oriented opinion words extraction (TOWE)~\cite{fan2019target} is a subtask of ABSA. Given a sentence and an aspect term occurring in the sentence, TOWE extracts the corresponding opinion words for the aspect term. For example, as shown in Figure~\ref{examples}, given the sentence ``Try the rose roll (not on menu).
'' and an aspect term ``rose roll'' appearing in the sentence, TOWE extracts the opinion word ``Try''. 

\begin{figure}[t]
	\centering
	\includegraphics[width=\linewidth]{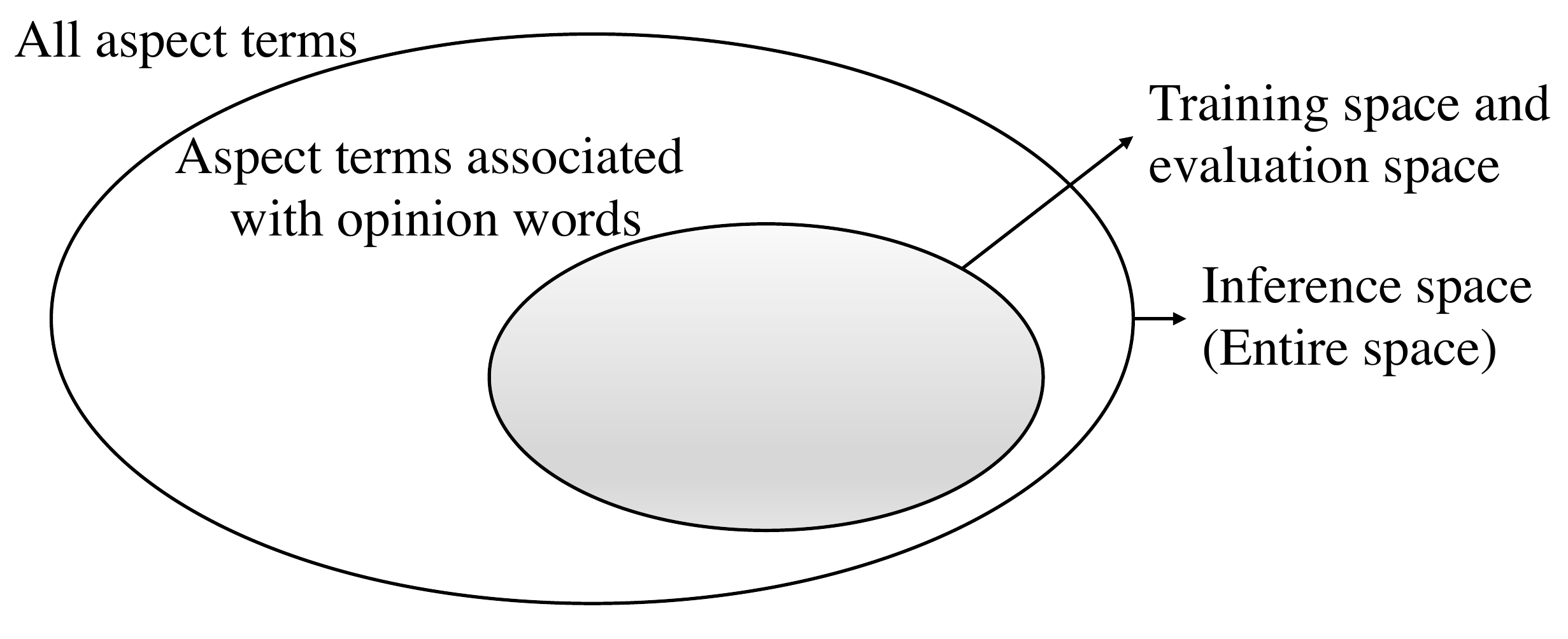}
	\caption{Illustration of sample selection bias problem in TOWE modeling. Training space and evaluation space are composed of aspect terms associated with at least one opinion word (i.e. Type I instances). It is only part of the inference space which is composed of all aspect terms. Note that inference happens in real-world scenarios. In real-world scenarios, TOWE models cannot extract opinion words only for Type I instances since they can not know whether an aspect term has opinion words in advance.}
	\label{sample-bias}
\end{figure}

TOWE has two types of instance. In the first type of instance, called \textbf{Type I instance}, aspect terms are associated with at least one opinion word. An example of Type I instance is shown in Figure~\ref{examples} (a). In the second type of instance, called \textbf{Type II instance}, aspect terms don't have corresponding opinion words. An example of Type II instance is shown in Figure~\ref{examples} (b). However, previous studies~\cite{fan2019target,wu2020latent,pouran-ben-veyseh-etal-2020-introducing,feng2021target,mensah-etal-2021-empirical,jiang-etal-2021-attention,kang2021rabert} trained and evaluated their models only on Type I instances and ignored Type II instances. The percentages of Type II instance in the four SemEval challenge datasets~\cite{pontiki-etal-2014-semeval, pontiki-etal-2015-semeval, pontiki-etal-2016-semeval}, which the four popular TOWE datasets only including Type I instances were built by~\citet{fan2019target} based on, range from 9.05\% to 32.33\%. This indicates that there is a considerable amount of Type II instances and Type II instances should not be ignored.

Furthermore, as illustrated in Figure~\ref{sample-bias}, ignoring Type II instances leads to a sample selection bias problem~\cite{zadrozny2004learning,ma2018entire}. Specifically, TOWE models are trained with only Type I instances, while these models will be utilized to make inference on the entire space with both Type I and Type II instances. Thus, the generalization performance of trained models will be hurt. Moreover, the performance of these models on Type I instances cannot reflect their performance on entire space, i.e. real-world scenarios.

To validate the sample selection bias problem, four popular TOWE datasets containing only Type I instances are extended and additionally include Type II instances. Experimental results on these datasets show that training TOWE models on entire space will significantly improve model performance and evaluating TOWE models only on Type I instances will overestimate model performance.

\section{Related Work}
Target-oriented opinion words extraction (TOWE) extracts the corresponding opinion words from sentences for a given aspect term and is proposed by \citet{fan2019target}. Moreover, \citet{fan2019target} built four TOWE datasets (i.e., Rest14, Lapt14, Rest15, and Rest16) based on four SemEval challenge datasets. The four SemEval challenge datasets include three restaurant datasets (i.e. Rest14, Rest15, and Rest16) from the SemEval Challenge 2014 Task 4~\cite{pontiki-etal-2014-semeval}, SemEval Challenge 2015 task 12~\cite{pontiki-etal-2015-semeval}, and SemEval Challenge 2016 task 5~\cite{pontiki-etal-2016-semeval}, and a laptop dataset (i.e. Lapt14) from the SemEval Challenge 2014 Task 4. In the original SemEval challenge datasets, the aspect terms are annotated, but the opinion words and the correspondence with aspect terms are not provided. Thus \citet{fan2019target} annotated the corresponding opinion words for the annotated aspect terms. Note that, in the four TOWE datasets that \citet{fan2019target} built, only the sentences
that contain pairs of aspect terms and opinion words are kept and only the aspect terms associated with at least one opinion term are used as instances. \citet{fan2019target} also proposed an Inward-Outward LSTM with Global context (IOG) for TOWE. 

Later, several other models were proposed for TOWE. \citet{pouran-ben-veyseh-etal-2020-introducing} proposed ONG including Ordered-Neuron Long Short-Term Memory Networks as well as GCN, and \citet{jiang-etal-2021-attention} proposed a novel attention-based relational graph convolutional neural network (ARGCN), both of which exploited syntactic information over dependency graphs to improve model performance. \citet{feng2021target} proposed Target-Specified sequence labeling with Multi-head Self-Attention for TOWE. \citet{wu2020latent} leveraged latent opinions knowledge from resource-rich review sentiment classification datasets to improve TOWE task. \citet{kang2021rabert} concentrated on incorporating aspect term information into BERT. \citet{mensah-etal-2021-empirical} conducted an empirical study to examine the actual contribution of position embeddings. These models obtained better performance. However, all these studies, following \citet{fan2019target}, only used Type I instances to train and evaluate their models.

\begin{table}
	\caption{Statistics of our new TOWE datasets. Ratio stands for the numbers of Type II instances to all instances in the dataset.}
	\label{tab:datasets}
	\begin{tabular}{|ccc|c|c|}
		\hline
		\multicolumn{3}{|c|}{Datasets}                                                                                       & \begin{tabular}[c]{@{}c@{}}\#aspect \\      terms\end{tabular} & Ratio (\%)                  \\ \hline
		\multicolumn{1}{|c|}{\multirow{6}{*}{Rest14-e}} & \multicolumn{1}{c|}{\multirow{2}{*}{training}}   & Type I instance & 2138                                                           & \multirow{2}{*}{28.35} \\ \cline{3-4}
		\multicolumn{1}{|c|}{}                        & \multicolumn{1}{c|}{}                            & Type II instance & 846                                                            &                        \\ \cline{2-5} 
		\multicolumn{1}{|c|}{}                        & \multicolumn{1}{c|}{\multirow{2}{*}{validation}} & Type I instance & 500                                                            & \multirow{2}{*}{29.58} \\ \cline{3-4}
		\multicolumn{1}{|c|}{}                        & \multicolumn{1}{c|}{}                            & Type II instance & 210                                                            &                        \\ \cline{2-5} 
		\multicolumn{1}{|c|}{}                        & \multicolumn{1}{c|}{\multirow{2}{*}{test}}       & Type I instance & 865                                                            & \multirow{2}{*}{23.72} \\ \cline{3-4}
		\multicolumn{1}{|c|}{}                        & \multicolumn{1}{c|}{}                            & Type II instance & 269                                                            &                        \\ \hline
		\multicolumn{1}{|c|}{\multirow{6}{*}{Lapt14-e}} & \multicolumn{1}{c|}{\multirow{2}{*}{training}}   & Type I instance & 1304                                                           & \multirow{2}{*}{32.33} \\ \cline{3-4}
		\multicolumn{1}{|c|}{}                        & \multicolumn{1}{c|}{}                            & Type II instance & 623                                                            &                        \\ \cline{2-5} 
		\multicolumn{1}{|c|}{}                        & \multicolumn{1}{c|}{\multirow{2}{*}{validation}} & Type I instance & 305                                                            & \multirow{2}{*}{30.21} \\ \cline{3-4}
		\multicolumn{1}{|c|}{}                        & \multicolumn{1}{c|}{}                            & Type II instance & 132                                                            &                        \\ \cline{2-5} 
		\multicolumn{1}{|c|}{}                        & \multicolumn{1}{c|}{\multirow{2}{*}{test}}       & Type I instance & 480                                                            & \multirow{2}{*}{26.72} \\ \cline{3-4}
		\multicolumn{1}{|c|}{}                        & \multicolumn{1}{c|}{}                            & Type II instance & 175                                                            &                        \\ \hline
		\multicolumn{1}{|c|}{\multirow{6}{*}{Rest15-e}} & \multicolumn{1}{c|}{\multirow{2}{*}{training}}   & Type I instance & 864                                                            & \multirow{2}{*}{9.05}  \\ \cline{3-4}
		\multicolumn{1}{|c|}{}                        & \multicolumn{1}{c|}{}                            & Type II instance & 86                                                             &                        \\ \cline{2-5} 
		\multicolumn{1}{|c|}{}                        & \multicolumn{1}{c|}{\multirow{2}{*}{validation}} & Type I instance & 212                                                            & \multirow{2}{*}{14.86} \\ \cline{3-4}
		\multicolumn{1}{|c|}{}                        & \multicolumn{1}{c|}{}                            & Type II instance & 37                                                             &                        \\ \cline{2-5} 
		\multicolumn{1}{|c|}{}                        & \multicolumn{1}{c|}{\multirow{2}{*}{test}}       & Type I instance & 436                                                            & \multirow{2}{*}{19.56} \\ \cline{3-4}
		\multicolumn{1}{|c|}{}                        & \multicolumn{1}{c|}{}                            & Type II instance & 106                                                            &                        \\ \hline
		\multicolumn{1}{|c|}{\multirow{6}{*}{Rest16-e}} & \multicolumn{1}{c|}{\multirow{2}{*}{training}}   & Type I instance & 1218                                                           & \multirow{2}{*}{12.94} \\ \cline{3-4}
		\multicolumn{1}{|c|}{}                        & \multicolumn{1}{c|}{}                            & Type II instance & 181                                                            &                        \\ \cline{2-5} 
		\multicolumn{1}{|c|}{}                        & \multicolumn{1}{c|}{\multirow{2}{*}{validation}} & Type I instance & 289                                                            & \multirow{2}{*}{15.99} \\ \cline{3-4}
		\multicolumn{1}{|c|}{}                        & \multicolumn{1}{c|}{}                            & Type II instance & 55                                                             &                        \\ \cline{2-5} 
		\multicolumn{1}{|c|}{}                        & \multicolumn{1}{c|}{\multirow{2}{*}{test}}       & Type I instance & 456                                                            & \multirow{2}{*}{25.49} \\ \cline{3-4}
		\multicolumn{1}{|c|}{}                        & \multicolumn{1}{c|}{}                            & Type II instance & 156                                                            &                        \\ \hline
	\end{tabular}
\end{table}

\begin{table*}[h]
	\caption{Performance of the models evaluated on entire space and Type I instances. All models are trained on Type I instances. Gains indicate how much higher the performance of the model evaluated on Type I instances is than the performance of the model evaluated on entire space. The best F1 score on entire space is marked in bold and the best F1 score on Type I instances is underlined.}
	\label{tab:evaluation}
	\begin{tabular}{|c|c|ccc|ccc|ccc|ccc|}
		\hline
		\multirow{2}{*}{Method}         & \multirow{2}{*}{Test instance type} & \multicolumn{3}{c|}{Rest14-e}    & \multicolumn{3}{c|}{Lapt14-e}    & \multicolumn{3}{c|}{Rest15-e}    & \multicolumn{3}{c|}{Rest16-e}    \\ \cline{3-14} 
		&                                     & P     & R     & F1             & P     & R     & F1             & P     & R     & F1             & P     & R     & F1             \\ \hline
		\multirow{3}{*}{ARGCN}          & Entire space                        & 71.70 & 81.30 & 76.17          & 60.99 & 69.48 & 64.90          & 68.10 & 75.90 & 71.72          & 69.85 & 81.91 & 75.37          \\ \cline{2-14} 
		& Type I instance                  & 85.25 & 82.29 & 83.73          & 74.00 & 70.41 & 72.14          & 76.56 & 75.66 & 76.05          & 86.00 & 82.33 & 84.10          \\
		& (Gains (\%))                        & 18.89 & 1.22  & 9.93           & 21.32 & 1.33  & 11.15          & 12.43 & -0.32 & 6.05           & 23.11 & 0.51  & 11.58          \\ \hline
		\multirow{3}{*}{ARGCN$_{bert}$} & Entire space                        & 73.19 & 83.46 & 77.98          & 63.19 & 73.22 & 67.80          & 73.88 & 74.73 & 74.27          & 74.26 & 83.78 & \textbf{78.72} \\ \cline{2-14} 
		& Type I instance                   & 86.00 & 83.01 & 84.46          & 75.79 & 76.02 & 75.88          & 78.21 & 75.74 & 76.92          & 86.91 & 84.05 & 85.45          \\
		& (Gains (\%))                        & 17.50 & -0.54 & 8.31           & 19.95 & 3.83  & 11.92          & 5.87  & 1.35  & 3.58           & 17.03 & 0.32  & 8.54           \\ \hline
		\multirow{3}{*}{IOG}            & Entire space                        & 73.02 & 76.62 & 74.77          & 61.60 & 67.25 & 64.19          & 70.78 & 69.98 & 70.32          & 68.37 & 81.41 & 74.29          \\ \cline{2-14} 
		& Type I instance                   & 82.80 & 78.66 & 80.64          & 72.43 & 69.63 & 70.96          & 77.19 & 69.86 & 73.29          & 85.67 & 79.81 & 82.58          \\
		& (Gains (\%))                        & 13.39 & 2.66  & 7.85           & 17.57 & 3.54  & 10.55          & 9.05  & -0.17 & 4.23           & 25.30 & -1.97 & 11.15          \\ \hline
		\multirow{3}{*}{IOG$_{bert}$}   & Entire space                        & 73.38 & 85.92 & \textbf{79.14} & 60.41 & 80.32 & \textbf{68.92} & 71.03 & 81.70 & \textbf{75.96} & 69.69 & 90.31 & 78.62          \\ \cline{2-14} 
		& Type I instance                   & 86.50 & 85.81 & {\ul 86.13}    & 77.62 & 80.92 & {\ul 79.22}    & 78.88 & 81.70 & {\ul 80.24}    & 88.68 & 89.66 & {\ul 89.16}    \\
		& (Gains (\%))                        & 17.88 & -0.14 & 8.84           & 28.50 & 0.75  & 14.95          & 11.05 & 0.00  & 5.64           & 27.25 & -0.72 & 13.40          \\ \hline
	\end{tabular}
\end{table*}

\begin{table*}
	\caption{
		Performance of the models trained on Type I instances and entire space. All models are evaluated on entire space.	Gains indicate how much better the model trained on entire space is than the model trained on Type I instances. The best F1 score is marked in bold.}
	\label{tab:training}
	\begin{tabular}{|c|c|ccc|ccc|ccc|ccc|}
		\hline
		\multirow{2}{*}{Method}         & \multirow{2}{*}{\begin{tabular}[c]{@{}c@{}}Training-validation\\      instance type\end{tabular}} & \multicolumn{3}{c|}{Rest14-e}    & \multicolumn{3}{c|}{Lapt14-e}                     & \multicolumn{3}{c|}{Rest15-e}    & \multicolumn{3}{c|}{Rest16-e}    \\ \cline{3-14} 
		&                                                                                                   & P     & R     & F1             & P     & \multicolumn{1}{c|}{R} & F1             & P     & R     & F1             & P     & R     & F1             \\ \hline
		\multirow{3}{*}{ARGCN}          & Type I instance                                                                                 & 71.70 & 81.30 & 76.17          & 60.99 & 69.48                  & 64.90          & 68.10 & 75.90 & 71.72          & 69.85 & 81.91 & 75.37          \\ \cline{2-14} 
		& Entire space                                                                                     & 81.02 & 78.29 & 79.58          & 73.29 & 64.94                  & 68.82          & 75.12 & 73.14 & 74.11          & 75.61 & 81.91 & 78.61          \\
		& (Gains (\%))                                                                                      & 13.00 & -3.70 & 4.48           & 20.16 & -6.54                  & 6.04           & 10.31 & -3.63 & 3.33           & 8.23  & 0.00  & 4.29           \\ \hline
		\multirow{3}{*}{ARGCN$_{bert}$} & Type I instance                                                                                 & 73.19 & 83.46 & 77.98          & 63.19 & 73.22                  & 67.80          & 73.88 & 74.73 & 74.27          & 74.26 & 83.78 & 78.72          \\ \cline{2-14} 
		& Entire space                                                                                     & 81.37 & 78.39 & 79.84          & 72.46 & 66.43                  & 69.27          & 76.63 & 72.50 & 74.48          & 78.68 & 81.03 & 79.82          \\
		& (Gains (\%))                                                                                      & 11.17 & -6.08 & 2.39           & 14.68 & -9.27                  & 2.17           & 3.73  & -2.98 & 0.29           & 5.96  & -3.28 & 1.39           \\ \hline
		\multirow{3}{*}{IOG}            & Type I instance                                                                                   & 73.02 & 76.62 & 74.77          & 61.60 & 67.25                  & 64.19          & 70.78 & 69.98 & 70.32          & 68.37 & 81.41 & 74.29          \\ \cline{2-14} 
		& Entire space                                                                                     & 75.78 & 74.68 & 75.19          & 71.08 & 62.91                  & 66.71          & 75.81 & 67.38 & 71.29          & 75.66 & 77.75 & 76.68          \\
		& (Gains (\%))                                                                                      & 3.78  & -2.53 & 0.56           & 15.39 & -6.44                  & 3.93           & 7.10  & -3.71 & 1.38           & 10.66 & -4.50 & 3.21           \\ \hline
		\multirow{3}{*}{IOG$_{bert}$}   & Type I instance                                                                                   & 73.38 & 85.92 & 79.14          & 60.41 & 80.32                  & 68.92          & 71.03 & 81.70 & 75.96          & 69.69 & 90.31 & 78.62          \\ \cline{2-14} 
		& Entire space                                                                                     & 81.64 & 80.87 & \textbf{81.24} & 71.64 & 75.24                  & \textbf{73.35} & 77.82 & 76.35 & \textbf{76.99} & 75.19 & 87.75 & \textbf{80.97} \\
		& (Gains (\%))                                                                                      & 11.26 & -5.88 & 2.66           & 18.60 & -6.32                  & 6.43           & 9.56  & -6.55 & 1.37           & 7.89  & -2.83 & 3.00           \\ \hline
	\end{tabular}
\end{table*}

\section{Experimental Setup}
\subsection{Datasets and Metrics}
To validate the sample selection bias problem in TOWE modeling, we built four new TOWE datasets (i.e. Rest14-e, Lapt14-e, Rest15-e, and Rest16-e) containing both Type I and Type II instances based on four popular TOWE datasets (i.e. Rest14, Lapt14, Rest15, and Rest16)~\cite{fan2019target} only including Type I instances. In our dataset names, the letter e stands for entire space. Specifically, \citet{fan2019target} built the four TOWE datasets by annotating the corresponding opinion words for the annotated aspect terms in four SemEval challenge datasets~\cite{pontiki-etal-2014-semeval, pontiki-etal-2015-semeval, pontiki-etal-2016-semeval}. However, only the aspect terms associated with at least one opinion word were kept and used as TOWE instances by~\citet{fan2019target}. To build our new TOWE datasets, first, our new TOWE datasets include all instances in the TOWE datasets built by~\citet{fan2019target}. Then, the aspect terms in the original SemEval challenge datasets, which don't have corresponding opinion words and hence were excluded from the TOWE datasets built by~\citet{fan2019target}, are added as a part of our new TOWE datasets. The statistics of our new TOWE datasets are shown in Table~\ref{tab:datasets}.

Following previous works~\cite{fan2019target}, we adopt evaluation metrics: precision (P), recall (R), and F1-score (F1). An extraction is considered as correct only when the opinion words from the beginning to the end are all predicted exactly as the ground truth.

\begin{table*}
	\caption{
		Performance of IOG$_{bert}$ trained on Type I instances and entire space. IOG$_{bert}$ is evaluated on Type I instances. The best F1 score is marked in bold.}
	\label{tab:impact-of-e-on-type1}
	\begin{tabular}{|c|c|ccc|ccc|ccc|ccc|}
		\hline
		\multirow{2}{*}{Method}   & \multirow{2}{*}{\begin{tabular}[c]{@{}c@{}}Training-validation\\      instance type\end{tabular}} & \multicolumn{3}{c|}{Rest14-e}  & \multicolumn{3}{c|}{Lapt14-e}                       & \multicolumn{3}{c|}{Rest15-e}   & \multicolumn{3}{c|}{Rest16-e}  \\ \cline{3-14} 
		&                                                                                                   & P     & R     & F1             & P                          & R     & F1             & P      & R     & F1             & P     & R     & F1             \\ \hline
		\multirow{2}{*}{IOG$_{bert}$} & Type I instance                                                                                   & 86.50 & 85.81 & \textbf{86.13} & 77.62                      & 80.92 & \textbf{79.22} & 78.88  & 81.70 & \textbf{80.24} & 88.68 & 89.66 & \textbf{89.16} \\ \cline{2-14} 
		& Entire space                                                                                      & 88.37 & 80.79 & 84.40          & \multicolumn{1}{c}{80.23} & 75.73 & 77.90          & 82.295 & 76.06 & 79.01          & 89.84 & 88.32 & 89.06          \\ \hline
	\end{tabular}
\end{table*}

\begin{table*}
	\caption{Case study. Incorrect predictions are marked in red.}
	\label{tab:cases}
	\begin{tabular}{|c|l|c|c|c|c|}
		\hline
		Id                  & Sentence                                                                                                                                         & Aspect term  & \begin{tabular}[c]{@{}c@{}}Training-validation\\      instance type\end{tabular} & Prediction                              & Ground truth \\ \hline
		&                                                                                                                                                  & chef         & Type I instance                                                               & {\color[HTML]{FE0000} {[}"not"{]}}      & {[}{]}       \\ \cline{3-6} 
		\multirow{-2}{*}{1} & \multirow{-2}{*}{\begin{tabular}[c]{@{}l@{}}Even when the chef is not in the house,\\ the food and service are right on target .\end{tabular}}   & chef         & Entire space                                                                    & {[}{]}                                  & {[}{]}       \\ \hline
		&                                                                                                                                                  & orange donut & Type I instance                                                               & {\color[HTML]{FE0000} {[}"never"{]}}    & {[}{]}       \\ \cline{3-6} 
		\multirow{-2}{*}{2} & \multirow{-2}{*}{\begin{tabular}[c]{@{}l@{}}I never had an orange donut before \\ so I gave it a shot .\end{tabular}}                            & orange donut & Entire space                                                                    & {[}{]}                                  & {[}{]}       \\ \hline
		&                                                                                                                                                  & Entrees      & Type I instance                                                               & {\color[HTML]{FE0000} {[}"classics"{]}} & {[}{]}       \\ \cline{3-6} 
		\multirow{-2}{*}{3} & \multirow{-2}{*}{\begin{tabular}[c]{@{}l@{}}Entrees include   classics like lasagna, \\ fettuccine Alfredo and chicken parmigiana.\end{tabular}} & Entrees      & Entire space                                                                    & {[}{]}                                  & {[}{]}       \\ \hline
		&                                                                                                                                                  & rose roll    &                                                                                  & {[}"Try"{]}                             & {[}"Try"{]}  \\ \cline{3-3} \cline{5-6} 
		&                                                                                                                                                  & menu         & \multirow{-2}{*}{Type I instance}                                              & {\color[HTML]{FE0000} {[}"Try"{]}}      & {[}{]}       \\ \cline{3-6} 
		&                                                                                                                                                  & rose roll    &                                                                                  & {\color[HTML]{FE0000} {[}{]}}           & {[}"Try"{]}  \\ \cline{3-3} \cline{5-6} 
		\multirow{-4}{*}{4} & \multirow{-4}{*}{Try the rose roll (not on menu).}                                                                                               & menu         & \multirow{-2}{*}{Entire space}                                                  & {[}{]}                                  & {[}{]}       \\ \hline
	\end{tabular}
\end{table*}

\subsection{TOWE Models}
We run experiments based on four TOWE models:
\begin{itemize}
	\item \textbf{ARGCN}~\cite{jiang-etal-2021-attention} first incorporates aspect term information by combining word representations with corresponding category embeddings with respect to the target tag of words. Then an attention-based relational graph convolutional network is used to learn semantic and syntactic relevance between words simultaneously. Finally, BiLSTM is utilized to capture the sequential information. Then obtained word representations are used to predict word tags $\{O, B, I\}$.
	\item \textbf{ARGCN$_{bert}$}~\cite{jiang-etal-2021-attention} is the BERT~\cite{devlin2018bert} version of ARGCN. The last hidden states of the pre-trained BERT are adopted as word representations and BERT is fine-tuned jointly. 
	\item \textbf{IOG}~\cite{fan2019target} uses an Inward-Outward LSTM to pass aspect term information to the left context and the right context of the aspect term, and obtains the aspect term-specific word representations. Then, a Bi-LSTM takes the aspect term-specific word representations as input and outputs the global contextualized word representations. Finally, the combination of the aspect term-specific word representations and the global contextualized word representations is used to predict word tags $\{O, B, I\}$.
	\item \textbf{IOG$_{bert}$} is the BERT version of IOG. Specifically, the word embedding layer and Inward-Outward LSTM in IOG are replaced with BERT. Moreover, BERT takes ``[CLS] sentence [SEP] aspect term [SEP]'' as input.
\end{itemize}

We run all models for 5 times and report the average results on the test datasets. 

\section{Results}
\subsection{Evaluation on Entire Space}
To observe the difference of TOWE model performance on Type I instances and entire space, the four models are trained on Type I instances (Both the training set and validation set only include Type I instances) like previous studies~\cite{fan2019target}, then are evaluated on both Type I instances and entire space. Experimental results are shown in Table~\ref{tab:evaluation}. We can see that  all models across all four datasets obtain much better performance on Type I instances than on entire space in terms of F1 score. For example, The performance gains of the best model $IOG_{bert}$ on Rest14-e, Lapt14-e, Rest15-e and Rest16-e are 8.84\%, 14.75\%, 5.64\% and 13.40\%, respectively. The increase on Rest15-e is smallest, since the ratio of Type II instances in Rest15-e is smallest (Table~\ref{tab:datasets}). Whatever, evaluating TOWE models on Type I instances will overestimate model performance.

\subsection{Training on Entire Space}
In this section, the four models are trained in two settings: (1) both the training set and validation set only include Type I instances, and (2) both the training set and validation set include both Type I and Type II Instances (i.e. entire space). Then the trained models are evaluated on entire space. Experimental results are shown in Table~\ref{tab:training}. From Table~\ref{tab:training} we draw the following two conclusions. First, all models trained on entire space outperform them trained on Type I instances in terms of F1 score across all four datasets, indicating that training models on entire space can improve the generalization performance of trained models. Second, while models trained on entire space obtain better precision, models trained on Type I instances obtain better recall. The reason is that the additional instances in entire space, i.e. Type II instances, only contain aspect terms without corresponding opinion words and hence help TOWE models to exclude incorrect opinion words for aspect terms, but also exclude some correct opinion words.  

$IOG_{bert}$ trained on Type I instances and entire space is also evaluated on Type I instances. The results are shown in Table~\ref{tab:impact-of-e-on-type1}. We can see in Table~\ref{tab:impact-of-e-on-type1} that $IOG_{bert}$ trained on Type I instances obtains better performance than $IOG_{bert}$ trained on entire space. This indicates that it is necessary to design models which work well on entire space and we leave this for future research.

\subsection{Case Study}
To further understand the impact of Type II instances on TOWE models, we show the predictions of IOG$_{bert}$ (the best TOWE model in our experiments) trained on Type I instance and entire space on four sentences. The four sentences are from the test set of the Rest14-e dataset. All the first three sentences contain only one aspect term and all the three aspect terms are Type II instances. While IOG$_{bert}$ trained on entire space makes correct inferences on the three instances,  IOG$_{bert}$ trained on Type I instances erroneously extracts opinion words for the three aspect terms. In fact, the words that  IOG$_{bert}$ trained on Type I instances extracts are not opinion words. The reason is that the entire-space training set of Rest14-e additionally includes Type II instances, some of which are even similar to the instances appearing in the first three sentences. For example, the sentence ``food was delivered by a busboy, not waiter'' and  the aspect term ``waiter'' in the sentence is a Type II instance in the training set of the Rest14-e.  This instance may suggest that IOG$_{bert}$ extract nothing for the aspect term ``chef'' from the first sentence. Thus, it is essential to train TOWE models on entire space. From the predictions of the fourth sentence, we can see that IOG$_{bert}$ trained on Type I instances prefers to extract more opinion words, while  IOG$_{bert}$ trained on entire space prefers to extract less opinion words. This is a case that show why the TOWE models trained on entire space obtain lower Recall than them trained on Type I instances.

\section{CONCLUSION}
In this paper, we explore the sample selection bias problem in target-oriented opinion words extraction (TOWE) modeling. Specifically, we divide TOWE instances into two types: Type I instances where the aspect terms are associated with at least one opinion word and Type II instances where the aspect terms don't have opinion words. Previous studies only use Type I instances to train and evaluate their models, resulting in a sample selection bias problem. Training TOWE models only on Type I instances may hurt the generalization performance of TOWE models. Evaluating TOWE models only on Type I instances can't reflect the performance of TOWE models on real-world scenarios. To validate our hypotheses, we add Type II instances to previous TOWE datasets only including Type I instances. Experimental results on these datasets demonstrate that training TOWE models on entire space including both Type I instances and Type II instances will significantly improve model performance and evaluating TOWE models only on Type I instances will overestimate model performance.

Since the datasets used by aspect sentiment triplet extraction (ASTE) \cite{peng2020knowing} are constructed based on the TOWE datasets built by~\citet{fan2019target}, hence may also have the sample selection bias problem. In the future, we will explore the sample selection bias problem on ASTE.

\bibliographystyle{ACM-Reference-Format}
\bibliography{sample-base,anthology,custom}

\appendix

\end{document}